\documentclass[conference]{IEEEtran}
\IEEEoverridecommandlockouts

\usepackage{cite}
\usepackage{amsmath,amssymb,amsfonts}
\usepackage{graphicx}
\usepackage{textcomp}
\usepackage{xcolor}

\graphicspath{{./images/}}

\def\BibTeX{{\rm B\kern-.05em{\sc i\kern-.025em b}\kern-.08em
    T\kern-.1667em\lower.7ex\hbox{E}\kern-.125emX}}

\begin{document}

\title{A Cascaded Edge-Cloud Architecture for Automated Diabetic Retinopathy Screening}

\author{
\IEEEauthorblockN{1\textsuperscript{st} Nishi Doshi}
\IEEEauthorblockA{\textit{University of Southern California}\\
Los Angeles, USA \\
nishimit@usc.edu}
\and
\IEEEauthorblockN{2\textsuperscript{nd} Shrey Shah}
\IEEEauthorblockA{\textit{University of Southern California}\\
Los Angeles, USA \\
shrey@alumni.usc.edu}
}

\maketitle

\begin{abstract}
Diabetic Retinopathy (DR) is one of the leading causes of preventable blindness, and automated screening can help extend specialist capacity in resource-constrained clinical workflows. Cloud-based deep learning systems can provide strong grading performance, but they require image upload and reliable connectivity. We evaluate a two-tier edge-cloud cascade on the public APTOS $2019$ Blindness Detection dataset. Tier $1$ runs a lightweight MobileNetV3-small model locally to triage Referable DR (Classes $2$--$4$) versus Non-referable DR (Classes $0$--$1$). Tier $2$ runs a RETFound-DINOv2 model in the cloud for ordinal severity grading only on images flagged as referable by Tier $1$. On a stratified APTOS test split of $733$ images, Tier $1$ reaches $98.99\%$ sensitivity and $84.37\%$ specificity at a validation-tuned high-sensitivity threshold. The deployment-oriented safety-floor cascade forwards $49.52\%$ of test images to Tier $2$, reducing image-count cloud calls by $50.48\%$ relative to a cloud-only model. In the deployed $4$-class output space (Class $0$--$1$ / Class $2$ / Class $3$ / Class $4$), the safety-floor cascade obtains $77.49\%$ accuracy and $0.7938$ quadratic weighted kappa, while the cloud-only baseline obtains $80.76\%$ accuracy and $0.8184$ kappa. A Tier-$2$-override comparator improves accuracy to $80.49\%$ and kappa to $0.8167$, but downgrades $12$ true referable cases after Tier $1$ correctly forwarded them. We also report repeated stratified bootstrap variability, a prevalence-dependent cloud-call curve, raw-byte upload estimates, and CPU latency for the local tier.
\end{abstract}


\section{Introduction}

Diabetic Retinopathy (DR) is a microvascular complication of diabetes mellitus and a leading cause of preventable blindness in working-age adults. A $2021$ meta-analysis estimates DR prevalence at $22.27\%$ of diabetic adults and projects $161$ million people with DR by $2045$\cite{Teo2021GlobalPO}\cite{article-six}; the Global Burden of Disease $2020$ study attributes $3.3$ million blindness cases and $10.8$ million moderate-to-severe vision impairment cases to DR-related causes\cite{article-eight}. DR is graded on the $5$-stage International Clinical Diabetic Retinopathy (ICDR) scale; Classes $0$--$1$ are treated as non-referable and Classes $2$--$4$ as referable \cite{article-six}.

Manual grading is labor-intensive and requires ophthalmologic training that is often unavailable in many screening programs. With a projected $18$ million health-worker shortage by $2030$ in lower-income regions\cite{article-eleven}, automated triage can help extend screening capacity. One approach is to use a large cloud model, which works well in connected hospitals but is harder to use in low-connectivity or metered-data settings.

Cloud-only DR screening faces practical barriers: fundus images can be large ($5$--$20$ MB), metered data can add operating cost for small clinics \cite{article-twenty}, and connectivity failures can interrupt results, as reported in the Google-Thailand DR deployment \cite{article-ten}.

Small edge models avoid the network dependency but are weaker for detailed grading. We therefore evaluate a two-tier cascade: Tier $1$ performs local referable/non-referable triage, and Tier $2$ performs ordinal severity grading only for images forwarded by Tier $1$. The final output is reported in $4$ deployed classes.

This study is a retrospective algorithmic evaluation on APTOS $2019$, not a field deployment study. APTOS is useful because it is public, labeled, and commonly used for DR grading, but it is not a rural, regional, or low-connectivity benchmark. We therefore frame bandwidth, cost, and latency as byte-level and CPU estimates derived from the same APTOS test split rather than as evidence of real-world clinical deployment benefit, consistent with DR readiness frameworks that separate benchmark feasibility from program readiness \cite{doshi2026clinicalinfrastructure}.

The cascade exploits four asymmetries: binary triage is easier than severity grading; Tier $1$ avoids a network call while Tier $2$ requires upload; APTOS $2019$ contains $59.4\%$ non-referable images, so savings must be interpreted against prevalence; and a Tier-$1$ false negative is more harmful than a false positive because it prevents cloud grading. We therefore ask whether a lightweight local model can provide high-sensitivity APTOS triage, how RETFound-DINOv2 performs as the cloud grader, how much cloud use the cascade avoids in the deployed $4$-class space, and how alternative cascade policies, prevalence, upload bytes, CPU latency, and repeated APTOS resampling affect the deployment trade-off.

\section{Related Works}

Automated DR analysis progressed from early CNN-based systems to deeper CNN architectures such as ResNet and multi-channel CNNs \cite{PRATT2016200, inproceedings}.

The FDA-cleared IDx-DR system provides a clinical reference point, with reported sensitivity of $95\%$ and specificity of $91\%$\cite{KHAN2025192}. Recent high-capacity models also use transformer-based designs; Dual-SwinOrd reports $87.98\%$ accuracy and Quadratic weighted kappa of $0.937$ \cite{bioengineering13040374}, while RET-CLIP and PRETI extend retinal foundation modeling with report or metadata supervision \cite{du2024retclip,lee2025preti}.

Edge models target local inference under limited compute and connectivity. Convolutional Vision Transformer (CvT) combines local feature extraction of CNNs with the global context of transformers and achieves $92.5\%$ accuracy, $91.20\%$ sensitivity and $94.0\%$ specificity\cite{article2025Vit}.

Prior edge-cloud DR frameworks preprocess fundus images locally before upload, reporting up to $80\%$ lower data transmission \cite{edge-computing}. More broadly, collaborative-intelligence systems such as Neurosurgeon show that DNN computation can be divided between mobile edge devices and cloud infrastructure to trade local compute against latency and energy \cite{kang2017neurosurgeon}. Cascaded medical-imaging designs have also been used for lung-nodule segmentation and classification \cite{shrey2020lungnodules}. Our cascade instead uses the edge model for referable/non-referable triage and forwards only referable images for cloud grading.

\section{Dataset}
\label{sec1}

We train and evaluate on the public APTOS $2019$ Blindness Detection dataset\cite{aptos2019}, which contains $3{,}662$ fundus images with strong class imbalance toward No DR and Moderate DR. Image sizes range from $474 \times 358$ to $4288 \times 2848$. Table \ref{tab1} shows the class distribution. Unlike newer multimodal DR datasets for foundation-model evaluation \cite{tang2026mmrdr}, APTOS does not provide rurality, clinic connectivity, geography, or deployment-cost metadata, so it is used here as a public DR benchmark rather than as a regional healthcare dataset.

\begin{table}
\begin{center}
\begin{tabular}{|c|c|c|}
\hline
Class & Stage & APTOS $2019$ count\\
\hline
$0$ & No DR & $1805$\\
\hline
$1$ & Mild DR & $370$\\
\hline
$2$ & Moderate DR & $999$\\
\hline
$3$ & Severe DR & $193$\\
\hline
$4$ & Proliferative DR & $295$\\
\hline
Total & - & $3662$\\
\hline
\end{tabular}
\end{center}
\smallskip
\caption{Number of images per class in APTOS $2019$ Dataset}
\label{tab1}
\end{table}

For Tier $1$, classes $0$ and $1$ are mapped to non-referable DR, while classes $2$--$4$ are mapped to referable DR. Cascade reporting uses four deployed classes: Non-referable DR (Class $0$--$1$), Class $2$, Class $3$, and Class $4$. Cases stopped as non-referable do not receive immediate cloud grading and would follow the screening program's routine follow-up policy.

\section{Proposed Pipeline}
\label{intro}

\begin{figure}[t]
    \centering
    \includegraphics[width=\linewidth]{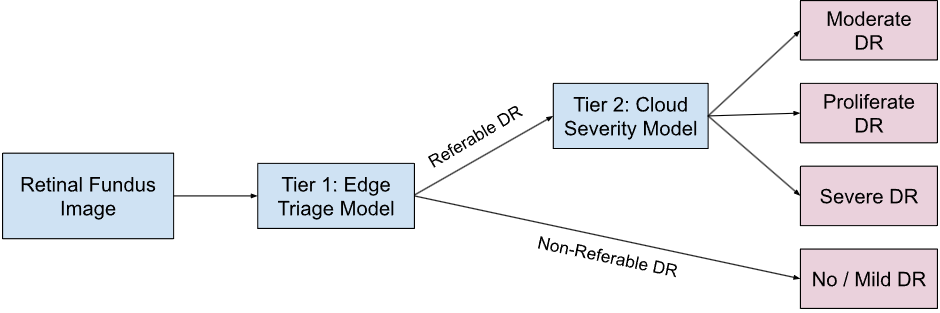}
    \caption{Two-tier edge-cloud cascade. A fundus image is first triaged at the edge; non-referable images exit as No/Mild DR, while referable images are sent to cloud Tier $2$ for Moderate, Severe, or Proliferative DR grading.}
    \label{fig:pipeline}
\end{figure}

We propose and evaluate a two-tier edge-cloud cascade (Figure~\ref{fig:pipeline}). Tier $1$ runs locally on MobileNetV3-small and performs binary triage between Referable DR (ICDR Classes $2$--$4$) and Non-referable DR (Classes $0$--$1$). As shown in Figure~\ref{fig:pipeline}, non-referable images bypass cloud grading and exit as the deployed No/Mild DR class, while referable images are forwarded to cloud Tier $2$, where RETFound-DINOv2 assigns Moderate, Severe, or Proliferative DR.

We distinguish two cascade policies. The deployment-oriented \textbf{safety-floor cascade} prevents a forwarded image from being downgraded below Class $2$, preserving the high-sensitivity Tier $1$ referral decision. The \textbf{Tier-$2$-override cascade} comparator lets Tier $2$ have final authority, including downgrading a forwarded image to non-referable, which tests whether the safety floor is worth its accuracy and kappa cost.

\subsection{Preprocessing}
\label{sec:preprocessing}
APTOS images vary substantially in size and contain dark borders around the retinal field of view. Tier $1$ uses the native ClementP MobileNetV3 preprocessing pipeline: red-channel foreground autocropping at threshold $10$, aspect-preserving resize, zero-padding to $512 \times 512$, and ImageNet normalization. Tier $2$ uses fundus cropping, bilinear resizing to $392 \times 392$, and ImageNet normalization for RETFound-DINOv2. Validation and test preprocessing is deterministic; no random augmentation is applied when reporting Tier $2$ or cascade metrics.

\subsection{Tier 1: Edge triage model}

Tier $1$ uses the released ClementP \texttt{FundusDRGrading-mobilenetv3\_small\_100} MobileNetV3-small checkpoint \cite{playout2024fundusdrgrading}, trained with $512 \times 512$ fundus images and a one-logit ordinal output head.

For our cascade, the ordinal output is passed through a sigmoid to produce a referable score, where higher values indicate stronger evidence for ICDR Classes $2$--$4$. The decision threshold is selected on the APTOS validation split using the validation-set sensitivity-specificity curve to prioritize sensitivity; the held-out test split is not consulted at any point during threshold selection. The selected operating point is
\begin{equation}
\sigma(z) \geq 0.6977,
\end{equation}
where $z$ is the model logit. We choose this sensitivity-biased threshold because a Tier $1$ false negative prevents cloud grading, while a false positive only adds one cloud inference. In the safety-floor policy, any forwarded image remains referable in the final deployed output even if Tier $2$ scores it below the Class $2$ boundary.

\subsection{Tier 2: Cloud severity model}

Tier $2$ uses RETFound-DINOv2 Vision Transformer Large with patch size $14$ (ViT-L/14) as the cloud-side severity model. It is a large retinal foundation model based on the RETFound framework \cite{zhou2023retfound}, with the DINOv2 \cite{oquab2024dinov2} self-supervised objective applied to color fundus photographs, making it a suitable initialization for DR grading. We do not claim this as a new model contribution; in this work, it is used as a pretrained backbone for the cloud stage of the cascade.

For APTOS, we attach a one-logit ordinal regression head and partially fine-tune the model on the training split. The model is trained using mean squared error (MSE) against the integer ICDR grade, so the output is a continuous severity score rather than a softmax probability over classes. At inference time, the score is rounded and clipped to the valid ICDR range $\{0,1,2,3,4\}$.

For standalone Tier $2$ reporting, we evaluate the model using the full $5$-class APTOS metrics. For comparison with the deployed cascade, Classes $0$ and $1$ are collapsed into a single non-referable class, giving the shared $4$-class output space: Class $0$-$1$, Class $2$, Class $3$, and Class $4$.

\subsection{Evaluation metrics}

We report Tier $1$ sensitivity, specificity, and accuracy for binary referable/non-referable triage, and report Tier $2$ and cascade accuracy plus quadratic weighted kappa ($\kappa$) for ordinal grading. Cascade metrics use the deployed four-class output space: Class $0$--$1$, Class $2$, Class $3$, and Class $4$. Cloud-call rate $X$ is the fraction of test images forwarded to Tier $2$, and image-count saving is $Y=100-X$. If referable prevalence is $p$, Tier $1$ sensitivity is $S$, and Tier $1$ specificity is $C$, the expected cloud-call rate is
\begin{equation}
    X(p) = S p + (1-C)(1-p).
\end{equation}
This is an image-count metric rather than a byte-level bandwidth measurement, since bandwidth also depends on resolution, compression, and transmission format.

\section{Experiment Setup}
\label{sec:experiment_setup}

We split APTOS $2019$ into stratified train, validation, and test sets using seed $42$: $64\%$ train, $16\%$ validation, and $20\%$ test. The same split is used for Tier $1$, Tier $2$, and the end-to-end cascade so the cloud-only and cascade results are compared on the same held-out test images.

\subsection{Tier $1$: edge triage model}
\label{sec:tier1_experiment}

Tier $1$ is evaluated as binary referable/non-referable triage using a sigmoid referable score from the MobileNetV3-small ordinal logit. The validation-selected threshold targets high sensitivity, and the fixed threshold is then evaluated on the held-out APTOS test split. CPU proxy latency is reported in Section \ref{sec:operational_results}.

\subsection{Tier $2$: cloud severity model}
\label{sec:tier2_experiment}

Tier $2$ is trained for $5$-class ordinal DR grading using RETFound-DINOv2 ViT-L/14 as the cloud-side backbone. We attach a $1024 \rightarrow 1$ ordinal head, optimize MSE against ICDR grade $\{0,1,2,3,4\}$, and round and clip outputs at inference. The last two transformer blocks, \texttt{fc\_norm}, and the head are trainable ($\approx25.2$M parameters); earlier layers are frozen. Optimization uses AdamW for $25$ epochs with mixed precision, weight decay $0.01$, backbone learning rate $1 \times 10^{-5}$, head learning rate $1 \times 10^{-3}$, and validation $5$-class QWK for checkpoint selection. For cascade comparison, Classes $0$ and $1$ are collapsed into the shared non-referable class.

\subsection{End-to-end cascaded pipeline}
\label{sec:e2e_cascade}

The full cascade is evaluated on the held-out APTOS test split in the same $4$-class output space as the cloud-only baseline. We report both the safety-floor cascade and the Tier-$2$-override comparator using the same Tier $1$ threshold and forwarded image set.

The main comparison is against the cloud-only Tier $2$ baseline, where every test image is processed by the RETFound-DINOv2 model. This comparison measures the trade-off introduced by the cascade: how much cloud use is reduced, and how much performance is retained in the deployed $4$-class output space.

\subsection{Repeated APTOS resampling and operational estimates}

The main models were trained once on the stratified seed-$42$ APTOS split. To quantify fixed-model variability, we perform $1{,}000$ stratified bootstrap resamples of the held-out test predictions, preserving the four deployed class counts; this is not a substitute for full repeated training splits or cross-validation. We also sum raw PNG bytes for the $733$ test images and the Tier-$1$-forwarded subset, and measure MobileNetV3-small CPU preprocessing plus inference on $128$ test images with batch size $1$, one PyTorch thread, and model loading excluded.

\section{Results}
\label{sec19}

\subsection{Tier $1$: edge triage results}
\label{sec:tier1_results}

At the validation-selected threshold, Tier $1$ correctly identifies $295$ of the $298$ referable test cases, yielding $98.99\%$ sensitivity, $90.31\%$ accuracy, and $84.37\%$ specificity. Thus the local triage model misses $3$ referable cases on the APTOS test split; this is retrospective test-set evidence, not clinical validation.

\subsection{Tier $2$: cloud severity results}
\label{sec:tier2_results}

In the standalone $5$-class ICDR output space, the RETFound-DINOv2 model achieves $79.54\%$ accuracy and a $5$-class QWK ($\kappa$) of $0.9101$.

For comparison with the deployed cascade, Classes $0$ and $1$ are collapsed into a single non-referable class, giving the shared $4$-class output space: Class $0$--$1$, Class $2$, Class $3$, and Class $4$. The cloud-only $4$-class accuracy is $80.76\%$.

The corresponding $4$-class QWK is $\kappa = 0.8184$. Because every test image is processed by Tier $2$ in this cloud-only baseline, the cloud-call rate is $100\%$.

\subsection{Cascaded pipeline results}
\label{sec:cascade_results}

\begin{figure*}[t]
    \centering
    \begin{minipage}{0.24\textwidth}
        \centering
        \includegraphics[width=\linewidth]{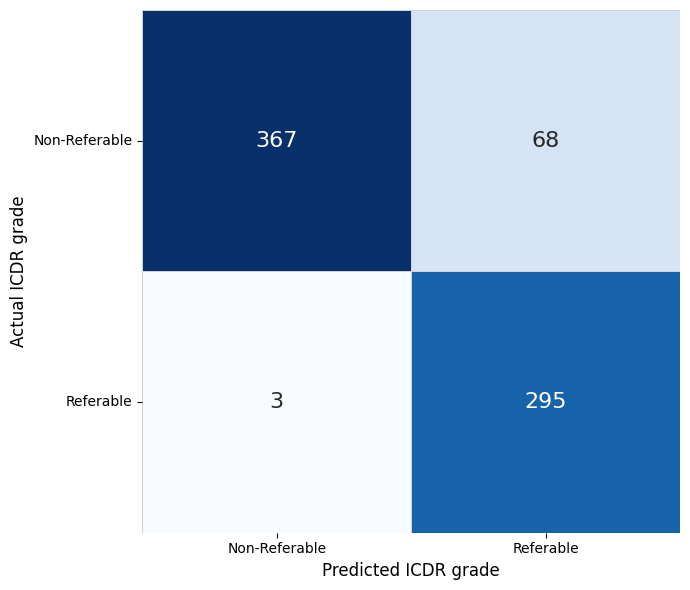}\\
        {\scriptsize (a) Tier $1$ triage}
    \end{minipage}
    \hfill
    \begin{minipage}{0.24\textwidth}
        \centering
        \includegraphics[width=\linewidth]{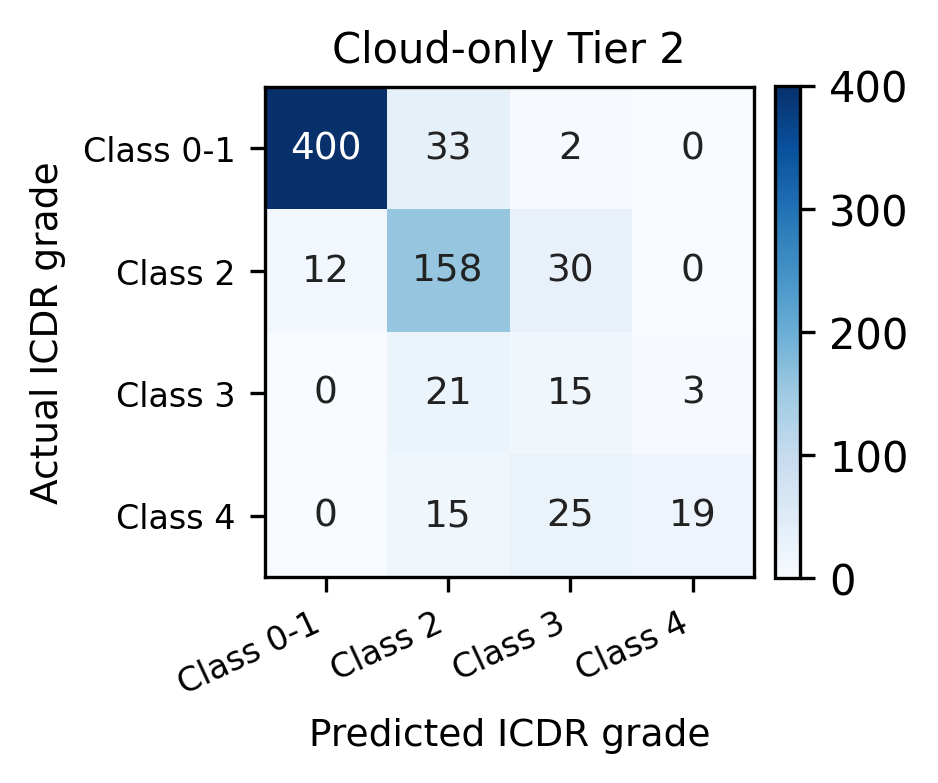}\\
        {\scriptsize (b) Cloud-only Tier $2$}
    \end{minipage}
    \hfill
    \begin{minipage}{0.24\textwidth}
        \centering
        \includegraphics[width=\linewidth]{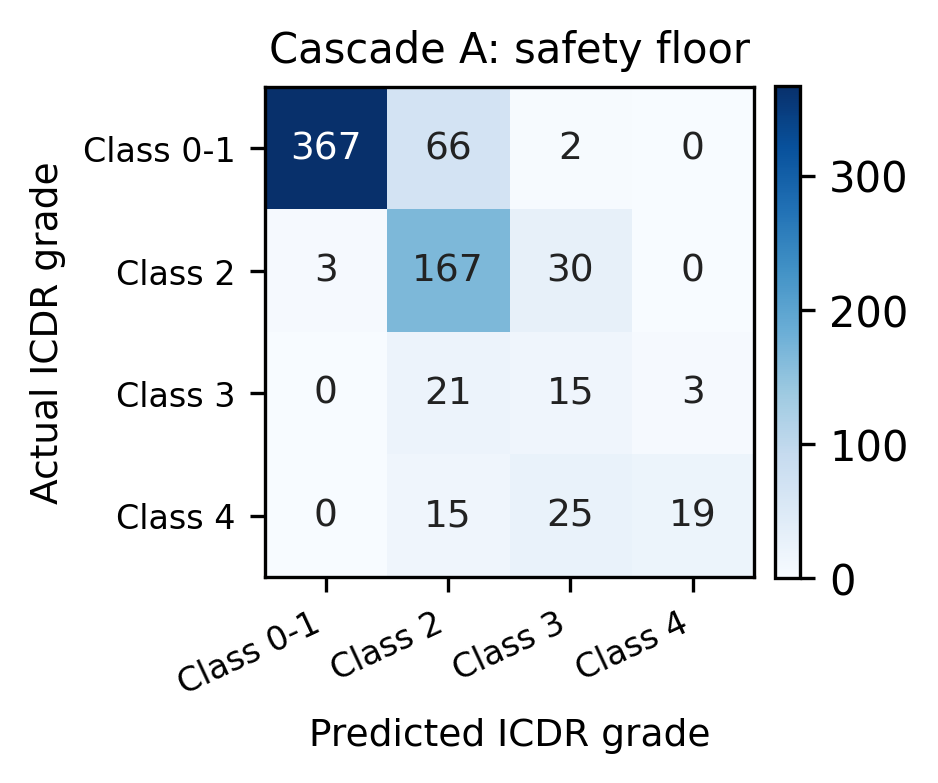}\\
        {\scriptsize (c) Safety-floor cascade}
    \end{minipage}
    \hfill
    \begin{minipage}{0.24\textwidth}
        \centering
        \includegraphics[width=\linewidth]{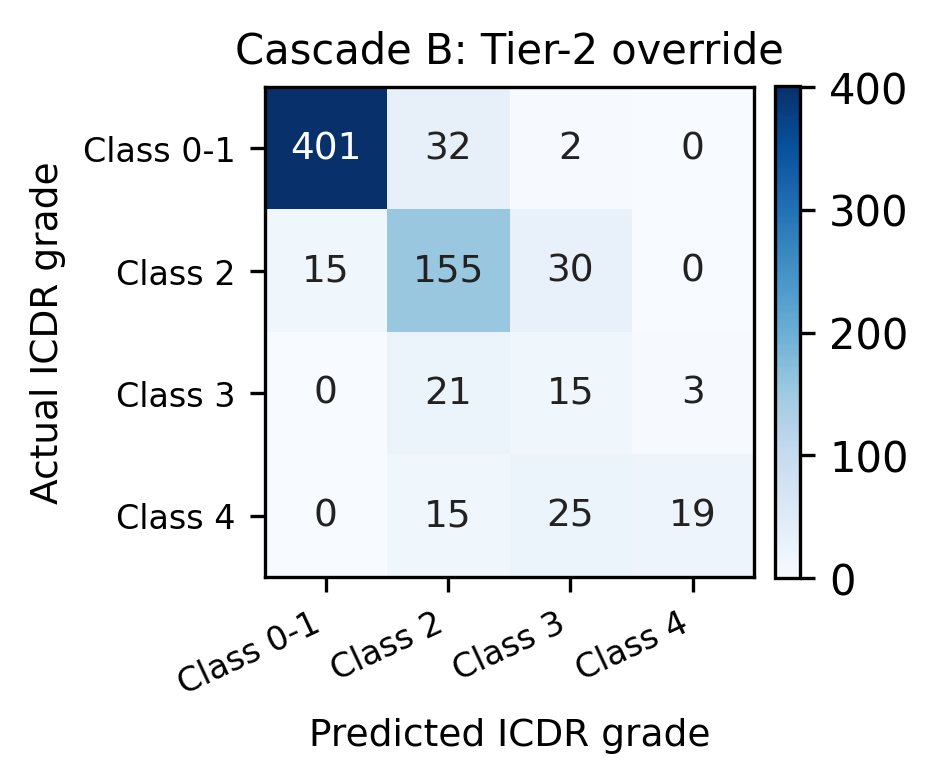}\\
        {\scriptsize (d) Tier-$2$ override}
    \end{minipage}
    \caption{Confusion matrices on the held-out APTOS test split. Rows are actual labels and columns are predicted labels.}
    \label{fig:confusion_matrices}
\end{figure*}

In both cascade policies, Tier $1$ acts as the forwarding gate: images with a referable score above the validation-selected threshold are sent to Tier $2$, while images below the threshold stop locally as non-referable. Figure~\ref{fig:confusion_matrices} shows the Tier $1$, cloud-only, and cascade confusion matrices.

For forwarded images, the safety-floor cascade uses Tier $2$ for severity grading but does not allow the final output to fall below Class $2$. The Tier-$2$-override comparator allows Tier $2$ to assign a forwarded image back to the non-referable Class $0$--$1$ group. This can correct false-positive Tier $1$ referrals, but it also introduces an additional safety-critical path in which true referable cases may be downgraded after being correctly forwarded.

Table~\ref{tab:cascade_comparison} compares the cloud-only baseline, the safety-floor cascade, and the Tier-$2$-override comparator. The safety-floor cascade forwards $49.52\%$ of test images and avoids cloud calls for the remaining $50.48\%$. Its deployed $4$-class accuracy is $77.49\%$ and its QWK is $\kappa=0.7938$. The Tier-$2$-override comparator uses the same forwarded image set but obtains $80.49\%$ accuracy and $\kappa=0.8167$.

The safety-floor policy prevents Tier $2$ from downgrading any true referable image that Tier $1$ correctly forwarded. Thus, the only final non-referable outputs among true referable cases are the $3$ Tier $1$ false negatives. Under the Tier-$2$-override comparator, $15$ true referable cases end as non-referable. Subtracting the $3$ Tier $1$ false negatives, $12$ true referable images were correctly forwarded by Tier $1$ but downgraded by Tier $2$.

\begin{table*}[h]
    \centering
    \small
    \begin{tabular}{|c|c|c|c|c|c|c|}
        \hline
         Model & Accuracy & $\kappa$ & Sensitivity & Specificity & Cloud-call rate ($X$) & Image-count saving ($Y$)\\
         \hline
         Cloud-only Tier $2$ baseline & $80.76\%$ & $0.8184$ & $95.97\%$ & $91.95\%$ & $100.00\%$ & $0.00\%$\\
         \hline
         Cascade A: safety floor & $77.49\%$ & $0.7938$ & $98.99\%$ & $84.37\%$ & $49.52\%$ & $50.48\%$\\
         \hline
          Cascade B: Tier-$2$ override & $80.49\%$ & $0.8167$ & $94.97\%$ & $92.18\%$ & $49.52\%$ & $50.48\%$\\
          \hline
    \end{tabular}
    \vspace{0.5em}
    \caption{Cloud-only baseline and two cascade policies in the deployed $4$-class output space. Cascade A is the deployment-oriented policy because it preserves Tier $1$ referral decisions. Cascade B is an ablation comparator that allows Tier $2$ downgrades.}
    \label{tab:cascade_comparison}
\end{table*}

This comparison highlights the main trade-off. The Tier-$2$-override policy retains nearly all cloud-only accuracy and QWK while reducing cloud calls by about half, but it lowers referable-case sensitivity from $98.99\%$ under the safety floor to $94.97\%$. The safety-floor policy is therefore the safer deployment choice when missed referral is weighted more heavily than exact severity grading.

\subsection{Prevalence, variability, and operational estimates}
\label{sec:operational_results}

\begin{figure}[h]
    \centering
    \includegraphics[width=0.98\linewidth]{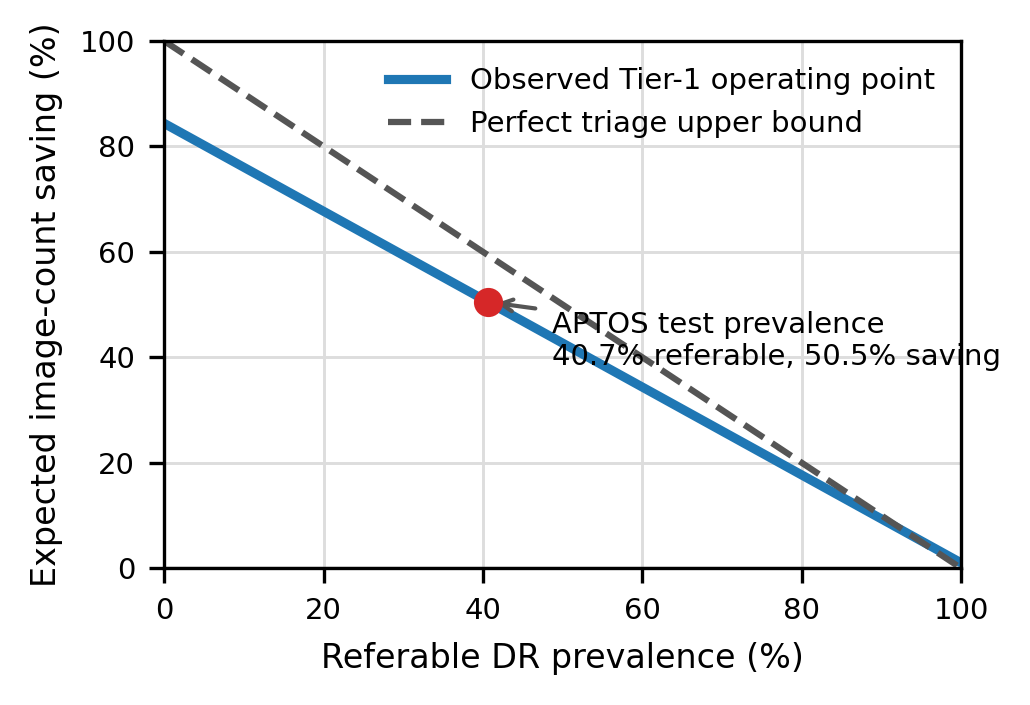}
    \caption{Expected image-count saving as a function of referable DR prevalence using the validation-selected Tier $1$ operating point.}
    \label{fig:prevalence_saving}
\end{figure}

The image-count saving depends on referable prevalence as well as Tier $1$ specificity. With Tier $1$ sensitivity $S=0.9899$ and specificity $C=0.8437$, the APTOS test split has referable prevalence $p=40.65\%$ and expected cloud-call rate $X(p)=49.52\%$. Figure~\ref{fig:prevalence_saving} plots the saving curve as prevalence changes. Thus, the APTOS saving is partly driven by the dataset's non-referable fraction, not only by a property discovered by the model.

Table~\ref{tab:bootstrap_variability} reports repeated stratified bootstrap variability over $1{,}000$ resamples of the held-out APTOS test predictions. The safety-floor cascade keeps high sensitivity with a mean of $99.01\% \pm 0.56\%$, while the Tier-$2$-override comparator has higher accuracy and QWK but lower sensitivity. These intervals describe resampling variability for the fixed trained models and should not be interpreted as full retraining cross-validation.

\begin{table*}[h]
    \centering
    \small
    \begin{tabular}{|c|c|c|c|c|}
        \hline
        Model & Accuracy & $\kappa$ & Sensitivity & Cloud-call rate\\
        \hline
        Cloud-only Tier $2$ baseline & $80.80\% \pm 1.26\%$ & $0.8183 \pm 0.0168$ & $95.94\% \pm 1.13\%$ & $100.00\% \pm 0.00\%$\\
        \hline
        Cascade A: safety floor & $77.53\% \pm 1.37\%$ & $0.7936 \pm 0.0178$ & $99.01\% \pm 0.56\%$ & $49.49\% \pm 1.05\%$\\
        \hline
        Cascade B: Tier-$2$ override & $80.52\% \pm 1.27\%$ & $0.8165 \pm 0.0167$ & $94.95\% \pm 1.22\%$ & $49.49\% \pm 1.05\%$\\
        \hline
    \end{tabular}
    \vspace{0.5em}
    \caption{Mean $\pm$ standard deviation over $1{,}000$ stratified bootstrap resamples of the held-out APTOS test predictions.}
    \label{tab:bootstrap_variability}
\end{table*}

For raw-byte transmission, the $733$ APTOS test PNGs occupy $1{,}718.0$ MB. The forwarded subset contains $363$ images and occupies $1{,}116.0$ MB, so the safety-floor cascade avoids $602.0$ MB of raw PNG upload on this test split. This is a $35.04\%$ raw-byte reduction, smaller than the $50.48\%$ image-count saving because forwarded referable images are larger on average ($3.07$ MB) than the overall test-set mean ($2.34$ MB). At a metered price of $C$ dollars per GB, the observed raw-byte saving is approximately $0.821C$ dollars per $1{,}000$ screened images.

On the local CPU timing run, Tier $1$ MobileNetV3-small with ClementP preprocessing has median end-to-end latency of $25.45$ ms per image and $95$th percentile latency of $28.99$ ms per image, measured with batch size $1$, one PyTorch CPU thread, and model loading excluded. This is a workstation CPU proxy rather than a dedicated edge-device benchmark, but it provides a concrete latency estimate for the local tier.

\section{Responsible AI and Limitations}
\label{sec:responsible_ai}

The principal harm in this cascade is a false non-referable output for a patient who actually has referable DR. In a screening workflow, this error can delay ophthalmology referral, treatment, and follow-up; for progressive disease, delay can increase the risk of preventable vision loss. The safety-floor cascade reduces one source of this harm by preventing Tier $2$ from downgrading a case that Tier $1$ has already referred. On the APTOS test split, this removes the $12$ Tier-$2$ downgrades observed in the override comparator. It does not remove Tier $1$ false negatives: $3$ of $298$ true referable APTOS test cases were stopped locally.

The system should therefore be treated as screening support rather than autonomous diagnosis. Practical safeguards include conservative thresholding, routine rescreening for non-referable outputs, human review of low-quality or uncertain images, fail-open referral when image quality or model confidence is inadequate, and local audit of false negatives after deployment. The threshold selected on APTOS may not transfer to populations with different cameras, image quality, disease prevalence, ethnicity, or care access. External validation and prospective monitoring are required before clinical use.

This work has several limitations. APTOS is not a rural or regional deployment dataset and does not include connectivity, clinical workflow, demographic, or cost metadata. Our repeated-resampling analysis quantifies uncertainty for fixed trained models, not variability from full retraining across folds. The byte analysis uses raw APTOS PNG files and does not model every possible upload format. The CPU timing is a local workstation proxy, not a benchmark on a dedicated clinic edge device. For this reason, the results are best viewed as retrospective feasibility evidence for an edge-cloud screening architecture, not as proof of real-world clinical or economic benefit.

\section{Conclusion}

This paper evaluated a two-tier edge-cloud cascade for automated DR screening on the public APTOS 2019 dataset. Tier $1$ uses a lightweight MobileNetV3-small model for referable/non-referable triage, while Tier $2$ uses RETFound-DINOv2 for ordinal severity grading on images forwarded by Tier $1$.

The Tier $1$ model achieved $98.99\%$ sensitivity on the held-out APTOS test split, missing $3$ of $298$ referable cases. In the deployment-oriented safety-floor cascade, the pipeline forwarded $49.52\%$ of test images to the cloud and avoided cloud calls for the remaining $50.48\%$. The safety-floor cascade achieved $77.49\%$ accuracy and $\kappa = 0.7938$ in the deployed $4$-class output space, compared with $80.76\%$ accuracy and $\kappa = 0.8184$ for the cloud-only Tier $2$ baseline. A Tier-$2$-override comparator improved accuracy to $80.49\%$ and $\kappa = 0.8167$, but it downgraded $12$ true referable cases after Tier $1$ correctly forwarded them.

These results suggest that a local triage stage can reduce cloud utilization substantially on APTOS, but the safety-performance trade-off depends on the cascade policy. The observed $50.48\%$ image-count saving corresponds to a $35.04\%$ raw-byte saving on the APTOS PNG files, and the local Tier $1$ CPU proxy runs with $25.45$ ms median preprocessing-plus-inference latency per image. The result should be interpreted as retrospective feasibility evidence, not clinical validation. Future work should evaluate the cascade on external datasets, perform full repeated training splits or cross-validation, benchmark real edge devices, and test the workflow in settings closer to resource-constrained screening practice.

\section*{Acknowledgment}
The authors acknowledge Cursor IDE and Claude AI agent assistance with code development and manuscript preparation.

\bibliographystyle{IEEEtran}
\bibliography{ref}

\end{document}